\documentclass{article}
\usepackage{spconf,amsmath,epsfig}
\usepackage[utf8]{inputenc} 
\usepackage[T1]{fontenc}    
\usepackage{hyperref}       
\usepackage{url}            
\usepackage{booktabs}       
\usepackage{amsfonts}       
\usepackage{amsmath}
\usepackage{amssymb}
\usepackage{nicefrac}       
\usepackage{microtype}      
\usepackage{xcolor}         

\usepackage{multirow}
\usepackage{adjustbox}

\title{Weak Labeling for Cropland Mapping in Africa}
\name{\begin{tabular}{c}
Gilles Quentin Hacheme\sthanks{Corresponding author: \texttt{ghacheme@microsoft.com}}\textsuperscript{1}, Akram Zaytar\textsuperscript{1}, Girmaw Abebe Tadesse\textsuperscript{1}, Caleb Robinson\textsuperscript{1}, Rahul Dodhia\textsuperscript{1} \\
Juan M. Lavista Ferres\textsuperscript{1}, Stephen Wood\textsuperscript{2}
\end{tabular}}

\address{Microsoft AI for Good Research Lab\textsuperscript{1}, The Nature Conservancy\textsuperscript{2}}

\begin{document}
%
\maketitle
\begin{abstract}
Cropland mapping can play a vital role in addressing environmental, agricultural, and food security challenges. However, in the context of Africa, practical applications are often hindered by the limited availability of high-resolution cropland maps. Such maps typically require extensive human labeling, thereby creating a scalability bottleneck. To address this, we propose an approach that utilizes unsupervised object clustering to refine existing weak labels, such as those obtained from global cropland maps. The refined labels, in conjunction with sparse human annotations, serve as training data for a semantic segmentation network designed to identify cropland areas. We conduct experiments to demonstrate the benefits of the improved weak labels generated by our method. In a scenario where we train our model with only 33 human-annotated labels, the $F_1$ score for the cropland category increases from $0.53$ to $0.84$ when we add the mined negative labels. 
\end{abstract}

\section{Introduction}
Up-to-date and high-resolution data on the spatial distribution of crop fields is critical for environmental, agricultural, and food security policies, especially in Africa, as most of the countries' economies heavily depend on agriculture \cite{diao2010role}. 
Cropland mapping from satellite imagery has been an essential topic due to its importance to derive data-driven insights and address climate and sustainability related challenges~\cite{potapov2022global, Kim_2021, adhikari2016evaluation, anderson2014comparative, boryan2011monitoring, santoro2017land}. However, most existing datasets only map croplands with low- to medium-sized resolution ($\geq30 m/pixel$ spatial resolution) from satellite imagery inputs such as Sentinel-2 or Landsat. Furthermore, it has been reported that existing land cover mapping solutions struggle to accurately map croplands in Africa~\cite{kerner2023accurate}. Specifically, Kerner et al. compare $11$ land cover datasets that cover Africa and contain a \textit{cropland} class and found that these maps have generally low levels of agreement compared to reference datasets from 8 countries on the continent. Locations with the highest agreement between maps are Mali ($69.9\%$) and Kenya ($60.6\%$) and the ones with the lowest agreement are Rwanda ($15.8\%$) and Malawi ($21.8\%$). If the goal is to achieve better results in specific regions, models that are tailored to those regions usually perform better than models that are designed for the whole world.

\begin{figure*}[th]
    \centering
    \includegraphics[width=0.9\textwidth]{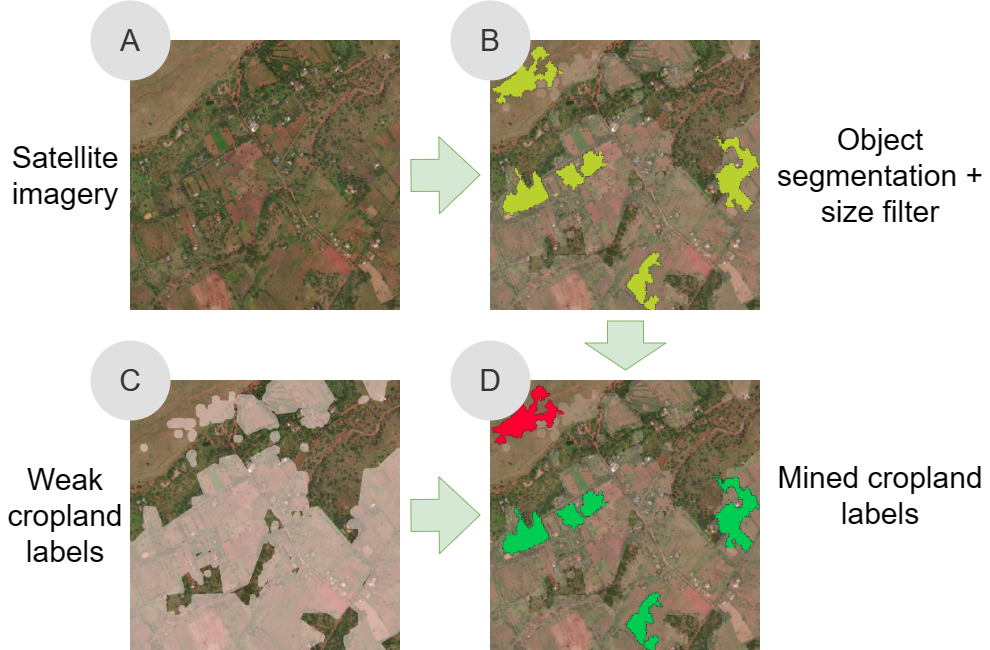}
    \caption{An overview of our proposed approach. Given satellite imagery \textbf{(A)} and weak cropland labels \textbf{(C)} over a given AOI we first use a K-Means clustering and filtering method to perform unsupervised object segmentation of the imagery \textbf{(B)}. We intersect the resulting objects (polygons) with the weak labels to mine stronger positive and negative samples \textbf{(D)}. Our experimental results show that adding these mined labels to human labels improves model performance.}
    \label{fig:overview}
\end{figure*}

To this end, we develop a modeling workflow for generating high-resolution cropland maps that are tailored toward a given area of interest (AOI), using Kenya as a use case.  We use a deep learning based semantic segmentation workflow -- an approach often employed for land-cover maps~\cite{robinson2021global, schmitt2020weakly, du2019smallholder, du2022dynamic, shuangpeng2019farmland}. In order to train the models, we used a mixture of sparse human labels gathered in the AOI and weak labels from global cropland maps. Specifically we use the area of intersection between an unsupervised object based clustering of the input satellite imagery and the weak labels to mine stronger \textit{cropland} (positive class) and \textit{non-cropland} (negative class) samples (see Figure~\ref{fig:overview} for an overview of this approach). We show that adding these labels to the human labels improves the $F_1$ score from $0.53$ to $0.84$ for the cropland class and $0.96$ to $0.99$ for the non-cropland class.

\section{Problem statement}\label{sec:problem_statement}

Consider a cropland mapping, i.e. semantic segmentation, problem over a given area of interest (AOI). We assume that we are given a large multi-spectral satellite image, a $k \times k$ dimensional matrix $A$ where $a_{ij}$ is the pixel from $A$ located at coordinates $(i,j)$. We also assume that we have a corresponding categorical mask $M^{\text{s}}$ with the same dimensions, derived from a human annotation of $A$, where each pixel, $m^{\text{s}}_{ij} \in \{0, 1, 2\}$, represents a class label, specifically $0=\textit{unknown}$, $1=\textit{non-cropland}$, $2=\textit{cropland}$. Note that the human annotation is often sparse, with only a few pixels annotated as either \textit{cropland} or \textit{non-cropland} and the majority of pixels are \textit{unknown}.  Further, we have an identically sized categorical mask $M^{\text{w}}$, derived from global cropland layers and/or from coarser resolution maps, where each each pixel, $m^{\text{w}}_{ij} \in \{1, 2\}$.  However, $M^{\text{w}}$ is assumed to have a higher level of label noise compared to  $M^{\text{s}}$.

In this work, we propose a data augmentation approach 
to generate an extended mask $M^{\text{e}}$, where $m^{\text{e}}_{ij} \in \{0, 1, 2\}$, that overcome the lack of strong \textit{cropland} and \textit{non-cropland} labels in $M^{\text{s}}$ by utilizing $M^{\text{w}}$.
In such a case, we hypothesize the semantic segmentation model should be improved by using the proposed data augmentation approach. 

\section{Cluster-based refinement of weak cropland labels}

\subsection{Data}
The AOI for the experiments in this paper is the \href{https://www.nature.org/en-us/what-we-do/our-insights/perspectives/foodscapes-regenerative-food-systems-nature-people/}{Central Highlands Ecoregion Foodscape} (CHEF) in Kenya.
We use Planetscope monthly basemap imagery, with spatial resolution of $4.7m/pixel$, provided by the \href{https://www.planet.com/nicfi/}{Norwegian International Climate and Forests Initiative} (NICFI) from January 2022 to December 2022.
We also use \href{https://maps.qed.ai/map/laikipia}{weak cropland labels} obtained from The Nature Conservancy (TNC) that cover the entire AOI. These labels do not delineate individual fields (i.e. when overlaid on Planet imagery the labels are not aligned with the imagery). This noise makes them insufficient for training a cropland segmentation model from high-resolution imagery (see Figure \ref{fig:overview}).
Finally, we manually annotate cropland and non-cropland areas by drawing polygons with respect to the high-resolution imagery. We avoid drawing large and coarse polygons to improve the delineation capability of our model.

\subsection{Method}
Our proposed method is to refine the weak labels by segmenting the high-resolution imagery, then intersecting each of the resulting objects (i.e. polygons) with the weak labels, and keeping objects that have high or low areas of intersection with the cropland class.

We first fit a K-means model on a subset of pixels randomly sampled from the $88$ imagery quads covering the CHEF region\footnote{We use $K=10$ clusters for this application based on visual validation, but this can differ for other applications.}. We randomly sample one million pixels out of the $4096\times4096 = 16,777,216$ pixels per quad, resulting in a sample size of $88$ million pixels, each with five features (one for each spectral band). Then we use the model to assign a cluster to each pixel in the original quad ($4096\times 4096$), save the predictions as a GeoTIFF, and again extract polygons from contiguous groups of pixels that are assigned to the same cluster (e.g. see Figure \ref{fig:overview}). We note that other unsupervised object based segmentation methods, such as the recently proposed Segment-Anything model~\cite{kirillov2023segment}, can be used in this step.

Next, we sequentially filter out polygons that are smaller than the $99^{th}$ quantile, then filter out remaining polygons that are larger than the $25^{th}$ quantile. This approach has been validated visually as the vast majority of the polygons are small and some polygons represents very large areas.

Finally, we estimate the proportion of cropland cover in each polygon by measuring the area of the polygon that intersects with the weak labels. The determination of cropland vs. non-cropland is then based on a threshold value of the intersection, $th$, measured in percentage. We classify a polygon as cropland when $th>80\%$ and non-cropland with $th<20\%$.
The result is enhanced weak labels that can be used to augment local strong labels for training a cropland semantic segmentation model.


\section{Experiments}\label{sec:sim}

To validate our proposed method, we run experiments where we consider training a cropland segmentation model under using different combinations of strong and weak labels within a single Planetscope scene (or quad)\footnote{\textit{L15-1237E-1025N}}. As our problem setting is to produce a \textit{map} of cropland areas in the specific AOI, without regards to generalization performance, we don't consider spatial or temporal generalization in our experimental setup and instead test on the same AOI. The scenarios considered in our experiments are as follows:

\begin{description}
    \item[Human labels:] We train the model on the AOI with the complete set of human labels, and we evaluate on the exact same AOI. This experiment is conducted for the sole reason of having the best performance level our system can potentially achieve given a more limited or noisier set of labels. In this experiment, we have 67 human labels (polygons) covering 4.056\% of the AOI.
    \item[Half human labels:] Here and in the following experiments we only use half of the human labels. This case is for simulating more realistic real-world scenarios where we only have a fraction of the whole data labeled by humans. 
    \item[Half human labels + mined labels:] This experiment extends the previous setting by adding \textit{all mined labels}, just the positive mined labels (\textit{mined positive labels}), or just the negative mined labels (\textit{mined negative labels}). 
    \item[Half human labels + weak labels:] Here we train the model with the half human label set and weak labels.
    \item [Half human labels + weak + mined negative labels:] Finally, we consider the case of training with the half human label set, weak labels, and the mined negative labels.
\end{description}

In all experiments our semantic segmentation model is the well-known U-Net~\cite{ronneberger2015u} with a ResNet-50 backbone~\cite{he2016deep}. It is trained using a cross-entropy loss function and the Adam optimization algorithm~\cite{kingma2014adam}. In each experiment we train the model using the given label set, then use the trained model to make predictions on the same imagery. The output produced by the model is a binary mask that shows the location of cropland regions in the input imagery.

\begin{table*}[t!]
\centering
\caption{Results derived from different scenarios of the labels considered. A detailed description of each experiment can be found in Section~\ref{sec:sim}.We report the number and area of mined labels for our proposed approach. We evaluate performance by measuring the $F_1$ score, Precision, and Pecall for each of the \textit{cropland} (C) and \textit{non-cropland} (NC) classes. We observe that adding mined negative labels to the human labels results in the best performance, improving significantly on only using human labels.}\label{tab:baseline}
\begin{adjustbox}{width=1.0\textwidth}
\begin{tabular}{lccccccc}
\toprule
\textbf{Scenario} & \textbf{Label} & \textbf{\begin{tabular}[c]{@{}c@{}}Mined\\ Labels (\#)\end{tabular}} & \textbf{\begin{tabular}[c]{@{}c@{}}Mined\\ Area (km$^2$)\end{tabular}} & $\mathbf{F_1}$ \textbf{Score} & \textbf{Precision} & \textbf{Recall} \\
\midrule
\multirow{2}{*}{Human labels} & C & - & - & 0.98 & 1.00 & 0.96 \\
& NC & - & - & 0.99 & 1.00 & 0.98 \\ 
\midrule 
\multirow{2}{*}{Half human labels} & C & - & - & 0.53 & 0.41 & 0.77 \\
& NC & - & - & 0.96 & 0.99 & 0.94 \\ \cmidrule{2-7}
\multirow{2}{*}{Half human labels + all mined labels} & C & 606 & 11.02 & 0.69 & 0.55 & 0.93 \\
& NC & 369 & 6.70 & 0.97 & 1.00 & 0.95 \\ \cmidrule{2-7}
\multirow{2}{*}{Half human labels + mined negative labels} & C & 0 & 0 & 0.84 & 0.92 & 0.78 \\
& NC & 369 & 6.70 & 0.99 & 0.99 & 0.98 \\ \cmidrule{2-7}
\multirow{2}{*}{Half human labels + mined positive labels} & C & 606 & 11.02 & 0.32 & 0.20 & 0.93 \\
& NC & 0 & 0 & 0.90 & 1.00 & 0.82 \\ \cmidrule{2-7}
\multirow{2}{*}{Half human labels + weak labels} & C & - & - & 0.29 & 0.17 & 0.96 \\
& NC & - & - & 0.88 & 1.00 & 0.79 \\ \cmidrule{2-7}
\multirow{2}{*}{Half human labels + weak labels + mined negative labels} & C & 0 & 0 & 0.58 & 0.42 & 0.96 \\
& NC & 369 & 6.70 & 0.96 & 1.00 & 0.93 \\ 
\bottomrule \\[-8px]
C = ``Cropland''; NC = ``Non-Cropland''
\end{tabular}
\end{adjustbox}
\end{table*}

Table \ref{tab:baseline} presents the performance of our semantic segmentation workflow for \textit{cropland} (C) and \textit{non-cropland} (NC) classes when experimented under different scenarios of labels (and their combinations) considered. The first experiment (\textit{Human labels}) leverages the complete set of human labels to simulate the ideal case. 
This experiment for \textit{cropland} achieves, as expected, a very high $F_1$ score of $0.98$, indicating overfitting of the model. The $F_1$ score for \textit{non-cropland} is even higher ($0.99$). 
These results are only helpful as they indicate results we could achieve if we had all the human labels at our disposal. But this scenario is usually less likely, and most of the time, we might get only a portion of the human labels. 

The following set of experiments shows results where only half the human labels are used in the training sets. The results show that as the number of human labels decreases (by half in this case), the $F_1$ scores globally decrease. The $F_1$ score for \textit{cropland} in the \textit{Half-human labels} experiment is only $0.53$, indicating a significant drop in performance compared to the ideal case. This drop is mainly due to a large decrease in the precision (only $0.41$). 
However, the performance for \textit{non-cropland} remains high, indicating that the segmentation task could still identify \textit{non-cropland} areas relatively well, even with fewer human labels. 
Using all the mined labels in addition to half the human labels (\textit{Half human labels + mined labels}) improves the cropland $F_1$ score from $0.53$ to $0.69$. But the highest $F_1$ score is obtained when only the negative mined samples are used in addition to half the human labels (\textit{Half human labels + mined negative labels}). The cropland $F_1$ score, in this case, reaches 0.84, with a precision of $0.92$, while the recall is almost the same as the one obtained with the \textit{Half human labels} experiment. 

Using the raw (positive) weak labels from TNC in addition to half the human labels (\textit{Half human labels + weak labels}), on the contrary, degrades the $F_1$ score for \textit{cropland} from $0.53$ to $0.29$. Even by combining the (positive) weak labels, the mined negative labels, and half human labels (\textit{Half human labels + weak labels + mined negative labels}), the $F_1$ score is only $0.58$. This confirms our assumption that the raw weak labels should not be used directly to augment the training set, and implicitly show the added value of our mining approach.

The key finding is that, in the scenario where we only use half the human labels in the training set, the $F_1$ score for the cropland category goes up from 0.53 to 0.84 when we include the mined negative labels.
This indicate the potential of mining weak labels for large-scale cropland mapping.

\section{Conclusion}
The accurate mapping of cropland fields through high-resolution satellite imagery is crucial for Africa's agricultural and food security policies. Unfortunately, the lack of high-quality cropland labels for Africa, e.g., clear delineation of farmlands, is the main bottleneck to exploit the growing capability of machine learning models to build high-resolution cropland maps. Unfortunately, models trained using cropland labels from other regions do not generalize well to unseen areas such as Africa. 
Our study presents a novel methodology to improve existing weak labels using K-means clustering, in order to augment existing training data, usually human labeled. The experimental results confirm that human labeling is vital for accurate results, while principled mining additional labels can significantly enhance large-scale cropland mapping. In a scenario where we train our model with only 50\% of the 67 human-annotated labels, adding the mined negative labels improves the $F_1$ score for the cropland category by almost 60\%.
Therefore, the proposed system could be an essential tool for large-scale cropland mapping. Future work includes validation of the proposed approach to multiple data sources and extended regions in Africa.

\bibliographystyle{IEEEbib}
\bibliography{strings,refs}

\end{document}